\documentclass{article} % For LaTeX2e
\usepackage{iclr2023_conference_tinypaper}
\usepackage{times}

\usepackage{amsmath,amsfonts,bm}

\def\eqref#1{equation~\ref{#1}}
\def\1{\bm{1}}

\DeclareMathAlphabet{\mathsfit}{\encodingdefault}{\sfdefault}{m}{sl}
\SetMathAlphabet{\mathsfit}{bold}{\encodingdefault}{\sfdefault}{bx}{n}

\usepackage{hyperref}
\usepackage{url}
\usepackage{graphicx}
\usepackage{subcaption}
\usepackage{booktabs}
\usepackage{multicol}

\title{On Difficulties of Attention Factorization through Shared Memory}

\author{Uladzislau Yorsh\textsuperscript{1}, Ond\v{r}ej Bojar\textsuperscript{1}, Martin Hole\v{n}a\textsuperscript{2}, David Herel\textsuperscript{2} \\
\textsuperscript{1}Charles University in Prague, \textsuperscript{2}Czech Technical University in Prague\\
\texttt{vlad.yorsh@mff.cuni.cz, hereldav@fel.cvut.cz}
}

\iclrfinalcopy % Uncomment for camera-ready version, but NOT for submission.
\begin{document}

\maketitle

\enlargethispage{\baselineskip}

%%Link to Code
\let\thefootnote\relax\footnotetext{Code: \url{https://github.com/vladyorsh/lra_efficient_transformers}}

%%This tiny negative spacing is needed to squeeze the code footnote in
\vspace{-0.125em}

\begin{abstract}
Transformers have revolutionized deep learning in numerous fields, including natural language processing, computer vision, and audio processing. Their strength lies in their attention mechanism, which allows for the discovering of complex input relationships. However, this mechanism's quadratic time and memory complexity poses challenges for larger inputs. Researchers are now investigating models like Linear Unified Nested Attention (Luna) or Memory Augmented Transformer, which leverage external learnable memory to either reduce the attention computation complexity down to linear, or to propagate information between chunks in chunk-wise processing. Our findings challenge the conventional thinking on these models, revealing that interfacing with the memory directly through an attention operation is suboptimal, and that the performance may be considerably improved by filtering the input signal before communicating with memory.

%Old version
%Researchers are now investigating models like Linear Unified Nested Attention (Luna), which uses external learnable memory to decrease attention computation to linear complexity. Our study focuses on Luna, particularly examining the influence of memory size and block connectivity on model training and accuracy. Our findings challenge conventional thinking about Luna, revealing that memory size has less effect on performance than anticipated, and that performance may be considerably enhanced by layer re-ordering, convolution layers, and max pooling.

\end{abstract}

\section{Introduction \& Related Work}

In the era of big data and natural language processing, handling long-form text is crucial. Transformers \citep{attention-is-all} have shown promise in some tasks, but they do not scale well with longer inputs due to their quadratic time and memory complexity inherent in their attention framework. This challenge has given rise to multiple approaches designed to handle sequences exceeding typical input lengths, including attention reformulation for efficient computing \citep{self-attention-does-not-need, flash}, exploration of weight sharing techniques \citep{universal-trans, t5}, heavy use of quantization \citep{qbert} or replacing the attention operation itself with a faster alternative.

In the present work, we focus on designs that alter the Transformer architecture to lower the computational demands by leveraging an external memory in the form of a set of learnable vectors. Models like Linear Unified Nested Attention (Luna; \citealp{luna}) or Perceiver \citep{perceiver} use it to factorize an attention operation into a sequence of attentions with a linear complexity, while the Memory Augmented Transformer \citep{memformer} processes long inputs chunk-by-chunk using the memory as a hidden state to carry information between chunks. While these models adopt different perspectives on long input processing, they all leverage the attention mechanism as an interface for communication between the input and memory. The latter can be used as a convenient fixed-length dense representation of sparse inputs such as texts.

Given the properties of the attention operation, we discover the phenomenon which does not allow to utilize multiple memory cells properly, which we call \textit{memory degradation}. Overcoming it may significantly improve the performance of the named models, and we propose several tweaks which lead to noticeable performance gains on the considered benchmarks.

\section{Methodology and Experimental Setup}

After inspecting the Luna and Set Transformer \citep{set-trans} memory states during training, we have found out that they tend to converge to a single or a small number of points. That means that the memory is not being used completely, and this conclusion is being supported by the experimental results from the Luna paper \citep{luna}, where the differences in performance between memory of sizes 16 and 256 are marginal. Unlike in vanilla Transformers, the attention logits of an input-memory attention matrix remain at relatively high entropy during training---the attention scores are distributed almost uniformly, and the resulting value vectors are similar for all input tokens. 

\begin{figure}[t]
    \captionsetup[subfigure]{font=scriptsize,labelfont=scriptsize}
    \centering
    %\vspace{-1cm}
    \begin{subfigure}[t]{0.24\textwidth}
         \centering
         \includegraphics[width=\textwidth]{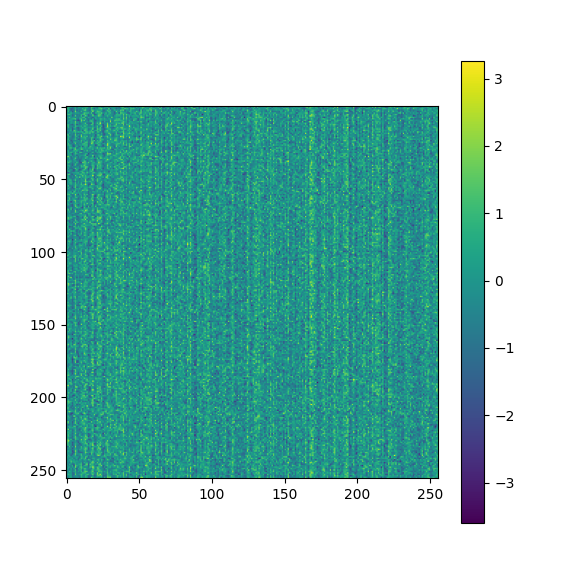}
         \caption{Luna memory}
     \end{subfigure}
     \hfill
     \begin{subfigure}[t]{0.24\textwidth}
         \centering
         \raisebox{2.3mm}{\includegraphics[width=\textwidth]{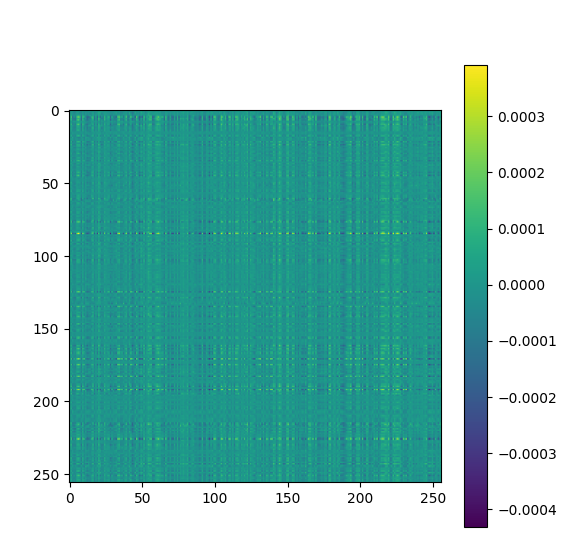}}
         \caption{Luna memory grad}
     \end{subfigure}
     \hfill
     \begin{subfigure}[t]{0.24\textwidth}
         \centering
         \includegraphics[width=\textwidth]{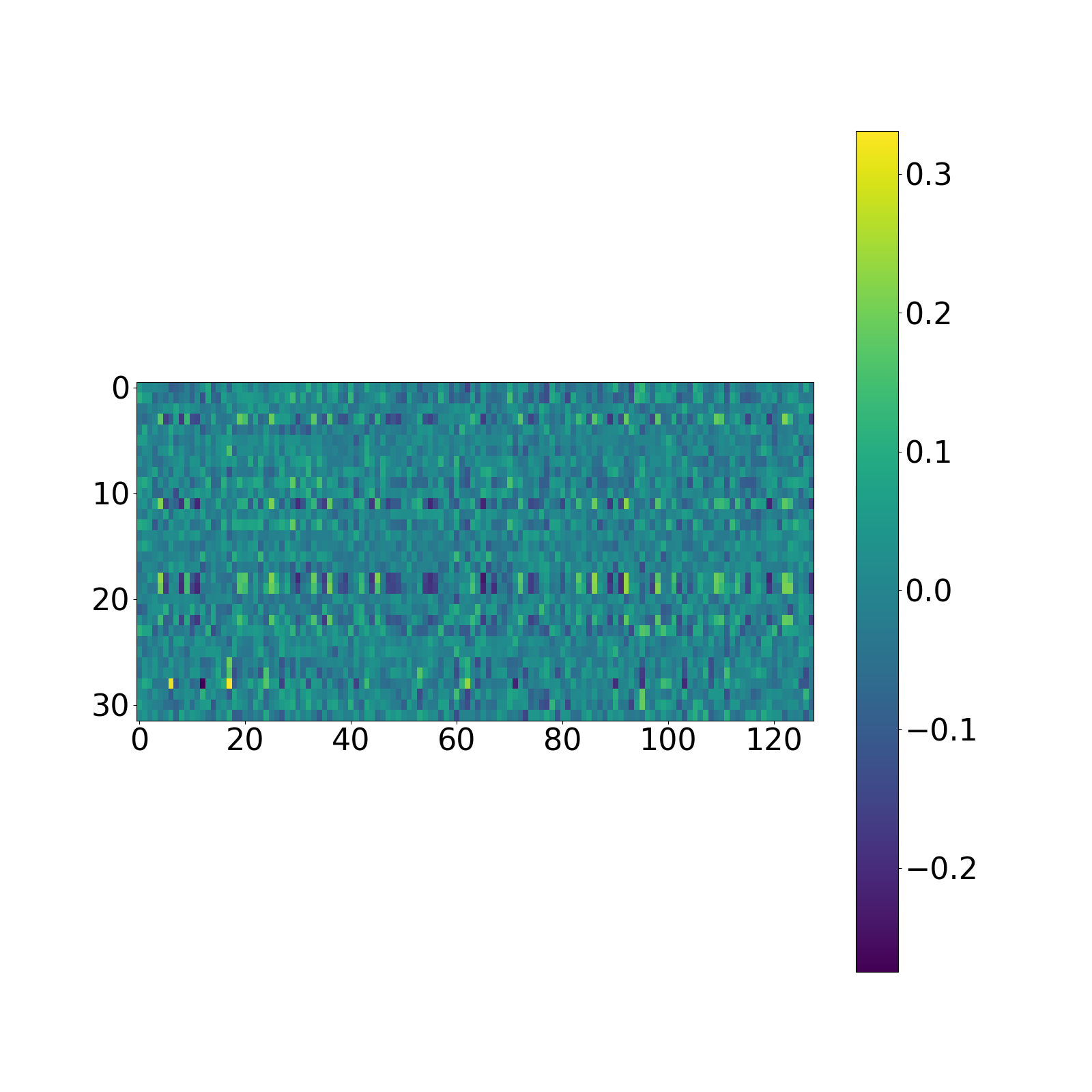}
         \caption{Set Transformer memory}
     \end{subfigure}
    \hfill
     \begin{subfigure}[t]{0.24\textwidth}
         \centering
         \includegraphics[width=\textwidth]{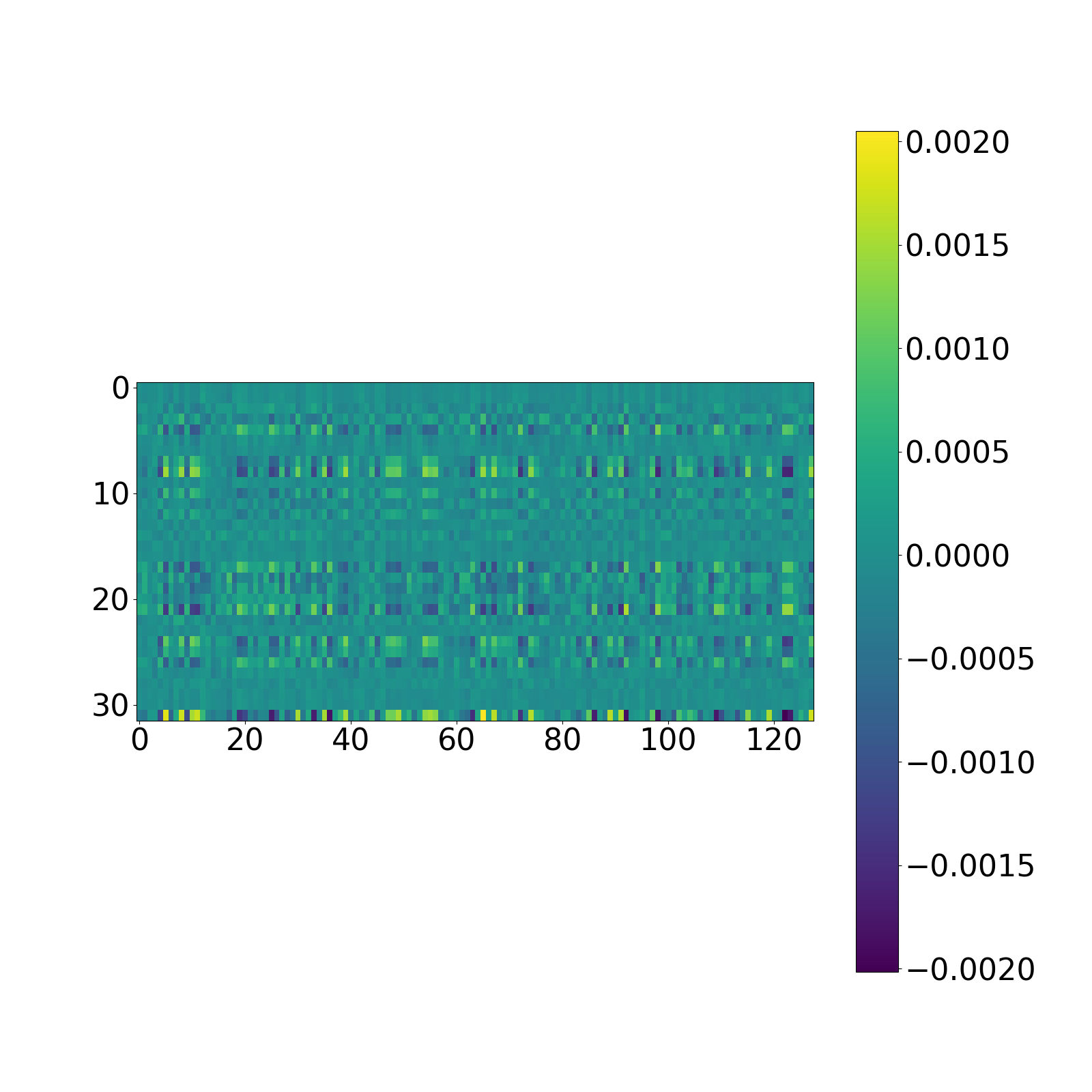}
         \caption{Set Transformer memory grad}
     \end{subfigure}
     \caption{Memory degradation illustrated. The horizontal axis is the feature dimension, the vertical one is the memory vector index, and the color indicates element values. Despite that memory matrices (learned parameters fed to the first model block as a memory input $P$ on the Figure~\ref{fig:modules}) were randomly initialized, during training they converge to a small number of unique vectors, see the vertical stripes on a) and b) and similar blocks on c) and d).}
\end{figure}

\enlargethispage{\baselineskip}

As an attempt to counter the issue of the memory degradation, we experiment with two techniques: filtering input before attending to the memory and lowering the softmax temperature (see Appendices~\ref{att-luna}~and~\ref{convluna_desc} for the detailed description). The first technique consists in applying a convolution or a pooling on the keys and values in the ``packing'' attention, leaving only the relevant signal which can form a better new memory representation after the attention. The second technique consists in replacing the $d^{-\frac{1}{2}}$ normalizing term in the attention equation with its learnable logarithm, initialized at zero. We apply both techniques only on ``packing'' attention, and refer to our model as ConvLuna.

We test our findings on the subset of the Long Range Arena (LRA; \citealp{lra}) benchmark. %, which consists of four classification tasks: binary sentiment analysis of BPE-encoded texts, BPE-encoded text matching task (predicting the citation link between two documents), ListOps (math operations over nested lists of digits) and CIFAR-10 (an image classification as a sequence of pixels). The benchmark suggests training from scratch and limiting the additional parametrization, which allows for a fairer comparison with other models.
We do not report the score for the Pathfinding task, because we have not managed to replicate the results of \citet{luna}. We provide a more detailed description of the experimental setup in Appendix~\ref{hyperparams} and an ablation study results in Appendix~\ref{ablation}. %To better see how memory is being utilized by the model, we use averaged memory cells instead of the [CLS] token as an input embedding before the final layer.

\begin{table}[b]
    \centering
    \begin{center}
    \resizebox{0.9\columnwidth}{!}{
    \begin{tabular}{llllll}
    \toprule
   Model &  Classification $\uparrow$ & Matching $\uparrow$ & ListOps $\uparrow$ & CIFAR-10 $\uparrow$ & Average $\uparrow$ \\
   % \\
    \midrule
     Transformer & 64.27 & 57.46 & 36.37 & 42.44 & 50.14 \\
    \midrule
    Luna-1 & 65.67 $\pm$ 0.18  & 75.46 $\pm$ 1.36 & 37.02 $\pm$ 0.12 & 49.06 $\pm$ 0.64 & 56.81\\
    Luna-16 & 65.53 $\pm$ 0.08 & 75.93 $\pm$ 0.89 & 36.98 $\pm$ 0.30 & 50.36 $\pm$ 0.43 & 57.20\\
    Luna-256 & 65.65 $\pm$ 0.35 & 79.44 $\pm$ 0.76 & 37.21 $\pm$ 0.22 & 50.90 $\pm$ 0.51 & 58.30 \\
    \midrule
     ConvLuna-1 & 82.10 $\pm$ 0.45 & \textbf{81.76} $\pm$ 0.73  & 43.95 $\pm$ 1.75 & 56.66 $\pm$ 0.44 & 66.12 \\
     ConvLuna-16 & \textbf{84.25} $\pm$ 0.16 & 80.47 $\pm$ 1.03 & \textbf{44.14} $\pm$ 0.69 & \textbf{56.93} $\pm$ 0.59 & \textbf{66.45} \\
     ConvLuna-256 & 83.29 $\pm$ 0.15 & 80.74 $\pm$ 1.08 &  43.56 $\pm$ 2.55 & 56.69 $\pm$ 0.64 & 65.90\\
    \bottomrule
    \end{tabular}
    }
    \end{center}
    \caption{Vanilla Transformer and Luna compared with ConvLuna. We report accuracy mean and standard deviation across five training runs for each setup. Values for the vanilla Transformer are taken from \citep{lra}, other results are by us. Numbers in names indicate memory size.
    \label{tab:lra-results}
    }
\end{table}

\section{Results and Conclusion}

% Table~\ref{tab:lra-results} indicates significant performance gains after an application of the proposed techniques. The improvements are especially noticeable on the classification ($\sim25\%$ of score) and CIFAR tasks. A peculiar result is that a single memory cell is enough to beat the vanilla Transformer on all the tasks, as well as many other baseline competitors \cite{lra}. However, the results also indicate that the problem is not yet completely resolved, and statistical tests (Appendix~\ref{stat-comp}) indicate there there is still no statistically significant influence of the memory size. We hypothesize that the observed improvements and persisting issues indicate a significant space for improvements, and the further work will be dedicated to finding a replacement for an attention as a memory-input interface.

%Our work demonstrates notable performance improvements in classification tasks and CIFAR datasets through our proposed methods, with even a single memory cell outperforming the standard Transformer model. These results highlight the inefficiency of traditional direct memory-attention interfacing and suggest that pre-filtering inputs before memory interaction can lead to significant gains. Although the impact of memory size on performance is not statistically significant, our findings indicate a substantial scope for future enhancements in Transformer efficiency and effectiveness.
Our work demonstrates notable performance improvements on several kinds of classification tasks through our proposed methods. We also find out that models with even a single memory cell outperform the standard Transformer model. These results highlight the inefficiency of the traditional direct input-memory interfacing through attention, and suggest that pre-filtering inputs before interacting with memory can lead to significant gains. However, throughout our experiments we were not able to achieve statistically significant impact of the memory size on the performance, which may indicate a substantial scope for future enhancements in efficiency and effectiveness of the architectures leveraging the external memory in a form of learnable vectors.

%\subsubsection*{Acknowledgements}

\subsubsection*{URM Statement}
The authors acknowledge that at least one key author of this work meets the URM criteria of ICLR 2024 Tiny Papers Track.

\enlargethispage{\baselineskip}

\subsubsection*{Acknowledgments}
Uladzislau Yorsh was supported by the grant №290223 of the Charles University Grant Agency. Ond\v{r}ej Bojar was supported by the grant 19-26934X (NEUREM3) of the
Czech Science Foundation.

\bibliography{iclr2023_conference_tinypaper}

\begin{thebibliography}{14}
\providecommand{\natexlab}[1]{#1}
\providecommand{\url}[1]{\texttt{#1}}
\expandafter\ifx\csname urlstyle\endcsname\relax
  \providecommand{\doi}[1]{doi: #1}\else
  \providecommand{\doi}{doi: \begingroup \urlstyle{rm}\Url}\fi

\bibitem[Dao et~al.(2022)Dao, Fu, Ermon, Rudra, and Ré]{flash}
Tri Dao, Daniel~Y. Fu, Stefano Ermon, Atri Rudra, and Christopher Ré.
\newblock Flashattention: Fast and memory-efficient exact attention with io-awareness, 2022.

\bibitem[Dehghani et~al.(2018)Dehghani, Gouws, Vinyals, Uszkoreit, and Kaiser]{universal-trans}
Mostafa Dehghani, Stephan Gouws, Oriol Vinyals, Jakob Uszkoreit, and Lukasz Kaiser.
\newblock Universal transformers.
\newblock \emph{CoRR}, abs/1807.03819, 2018.
\newblock URL \url{http://arxiv.org/abs/1807.03819}.

\bibitem[Garc{{\'i}}a \& Herrera(2008)Garc{{\'i}}a and Herrera]{garcia}
Salvador Garc{{\'i}}a and Francisco Herrera.
\newblock An extension on ``statistical comparisons of classifiers over multiple data sets'' for all pairwise comparisons.
\newblock \emph{Journal of Machine Learning Research}, 9\penalty0 (89):\penalty0 2677--2694, 2008.
\newblock URL \url{http://jmlr.org/papers/v9/garcia08a.html}.

\bibitem[Jaegle et~al.(2021)Jaegle, Gimeno, Brock, Zisserman, Vinyals, and Carreira]{perceiver}
Andrew Jaegle, Felix Gimeno, Andrew Brock, Andrew Zisserman, Oriol Vinyals, and Jo{\~{a}}o Carreira.
\newblock Perceiver: General perception with iterative attention.
\newblock \emph{CoRR}, abs/2103.03206, 2021.
\newblock URL \url{https://arxiv.org/abs/2103.03206}.

\bibitem[Krizhevsky(2009)]{cifar}
Alex Krizhevsky.
\newblock Learning multiple layers of features from tiny images.
\newblock Technical report, University of Toronto, Toronto, Ontario, 2009.
\newblock URL \url{https://www.cs.toronto.edu/~kriz/learning-features-2009-TR.pdf}.

\bibitem[Lee et~al.(2019)Lee, Lee, et~al.]{set-trans}
Juho Lee, Yoonho Lee, et~al.
\newblock Set transformer: A framework for attention-based permutation-invariant neural networks.
\newblock In \emph{Proceedings of the 36th International Conference on Machine Learning}, volume~97 of \emph{Proceedings of Machine Learning Research}, pp.\  3744--3753. PMLR, 09--15 Jun 2019.
\newblock URL \url{https://proceedings.mlr.press/v97/lee19d.html}.

\bibitem[Ma et~al.(2021)Ma, Kong, Wang, Zhou, May, Ma, and Zettlemoyer]{luna}
Xuezhe Ma, Xiang Kong, Sinong Wang, Chunting Zhou, Jonathan May, Hao Ma, and Luke Zettlemoyer.
\newblock Luna: Linear unified nested attention, 2021.

\bibitem[Rabe \& Staats(2021)Rabe and Staats]{self-attention-does-not-need}
Markus~N. Rabe and Charles Staats.
\newblock Self-attention does not need o(n\({}^{\mbox{2}}\)) memory.
\newblock \emph{CoRR}, abs/2112.05682, 2021.
\newblock URL \url{https://arxiv.org/abs/2112.05682}.

\bibitem[Radev et~al.(2009)Radev, Muthukrishnan, and Qazvinian]{acl}
Dragomir~R. Radev, Pradeep Muthukrishnan, and Vahed Qazvinian.
\newblock The {ACL} {A}nthology network corpus.
\newblock In \emph{Proceedings of the 2009 Workshop on Text and Citation Analysis for Scholarly Digital Libraries ({NLPIR}4{DL})}, pp.\  54--61, Suntec City, Singapore, August 2009. Association for Computational Linguistics.
\newblock URL \url{https://aclanthology.org/W09-3607}.

\bibitem[Raffel et~al.(2019)Raffel, Shazeer, Roberts, Lee, Narang, Matena, Zhou, Li, and Liu]{t5}
Colin Raffel, Noam Shazeer, Adam Roberts, Katherine Lee, Sharan Narang, Michael Matena, Yanqi Zhou, Wei Li, and Peter~J. Liu.
\newblock Exploring the limits of transfer learning with a unified text-to-text transformer.
\newblock \emph{CoRR}, abs/1910.10683, 2019.
\newblock URL \url{http://arxiv.org/abs/1910.10683}.

\bibitem[Shen et~al.(2019)Shen, Dong, et~al.]{qbert}
Sheng Shen, Zhen Dong, et~al.
\newblock {Q-BERT:} hessian based ultra low precision quantization of {BERT}.
\newblock \emph{CoRR}, abs/1909.05840, 2019.
\newblock URL \url{http://arxiv.org/abs/1909.05840}.

\bibitem[Tay et~al.(2020)Tay, Dehghani, Abnar, Shen, Bahri, Pham, Rao, Yang, Ruder, and Metzler]{lra}
Yi~Tay, Mostafa Dehghani, Samira Abnar, Yikang Shen, Dara Bahri, Philip Pham, Jinfeng Rao, Liu Yang, Sebastian Ruder, and Donald Metzler.
\newblock Long range arena: {A} benchmark for efficient transformers.
\newblock \emph{CoRR}, abs/2011.04006, 2020.
\newblock URL \url{https://arxiv.org/abs/2011.04006}.

\bibitem[Vaswani et~al.(2017)Vaswani, Shazeer, et~al.]{attention-is-all}
Ashish Vaswani, Noam Shazeer, et~al.
\newblock Attention is all you need.
\newblock In \emph{Advances in Neural Information Processing Systems}, volume~30. Curran Associates, Inc., 2017.
\newblock URL \url{https://proceedings.neurips.cc/paper_files/paper/2017/file/3f5ee243547dee91fbd053c1c4a845aa-Paper.pdf}.

\bibitem[Wu et~al.(2022)Wu, Lan, et~al.]{memformer}
Qingyang Wu, Zhenzhong Lan, et~al.
\newblock Memformer: A memory-augmented transformer for sequence modeling.
\newblock In \emph{Findings of the Association for Computational Linguistics: AACL-IJCNLP 2022}, pp.\  308--318, Online only, November 2022. Association for Computational Linguistics.
\newblock URL \url{https://aclanthology.org/2022.findings-aacl.29}.

\end{thebibliography}
\bibliographystyle{iclr2023_conference_tinypaper}

\newpage
\appendix

\section{Statistical Comparison of Different Memory Sizes}
\label{stat-comp}

\begin{figure}[h]
    \centering
    \includegraphics[width=0.75\linewidth]{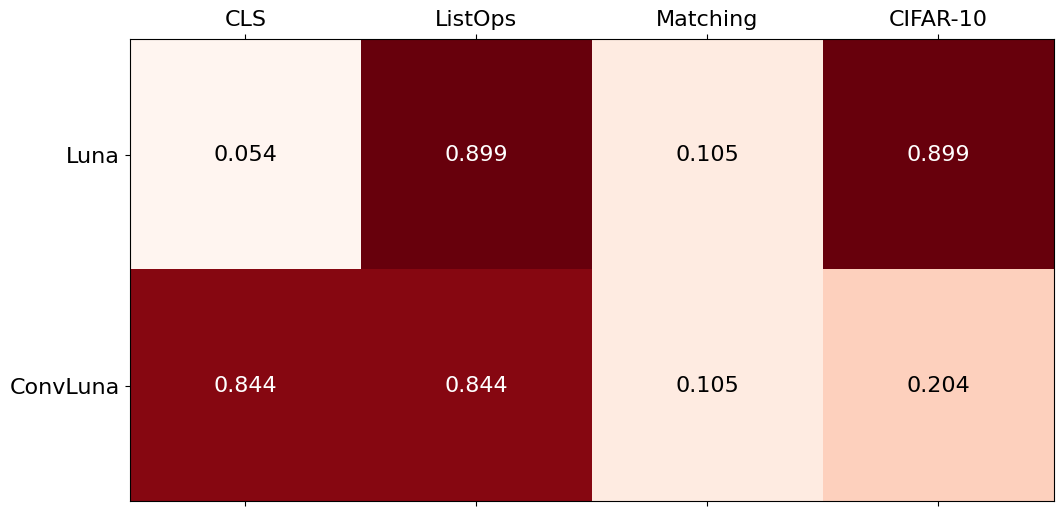}
    \caption{Achieved significances (p-values) of the Friedman test across all considered memory sizes with the H0 = ``expected accuracies are equal''. Color codes and numbers correspond to the p-values, corrected using the Holm metod \citep{garcia}. We could not reject the null hypothesis on the 5\% level of significance for any setup.}
    \label{fig:friedman}
\end{figure}

To find out whether the increasing memory actually provides statistically significant performance gains, we conduct the Friedman test (Figure~\ref{fig:friedman}) over accuracy score samples for memory sizes \{1, 16, 256\} for each combination of the model and the task. As mentioned before, each sample thus contains five experiments.

\section{Attention and Luna Definition}
\label{att-luna}

\begin{figure}[ht]
    \centering
     \begin{subfigure}[b]{0.32\textwidth}
         \centering
         \includegraphics[width=0.53\textwidth]{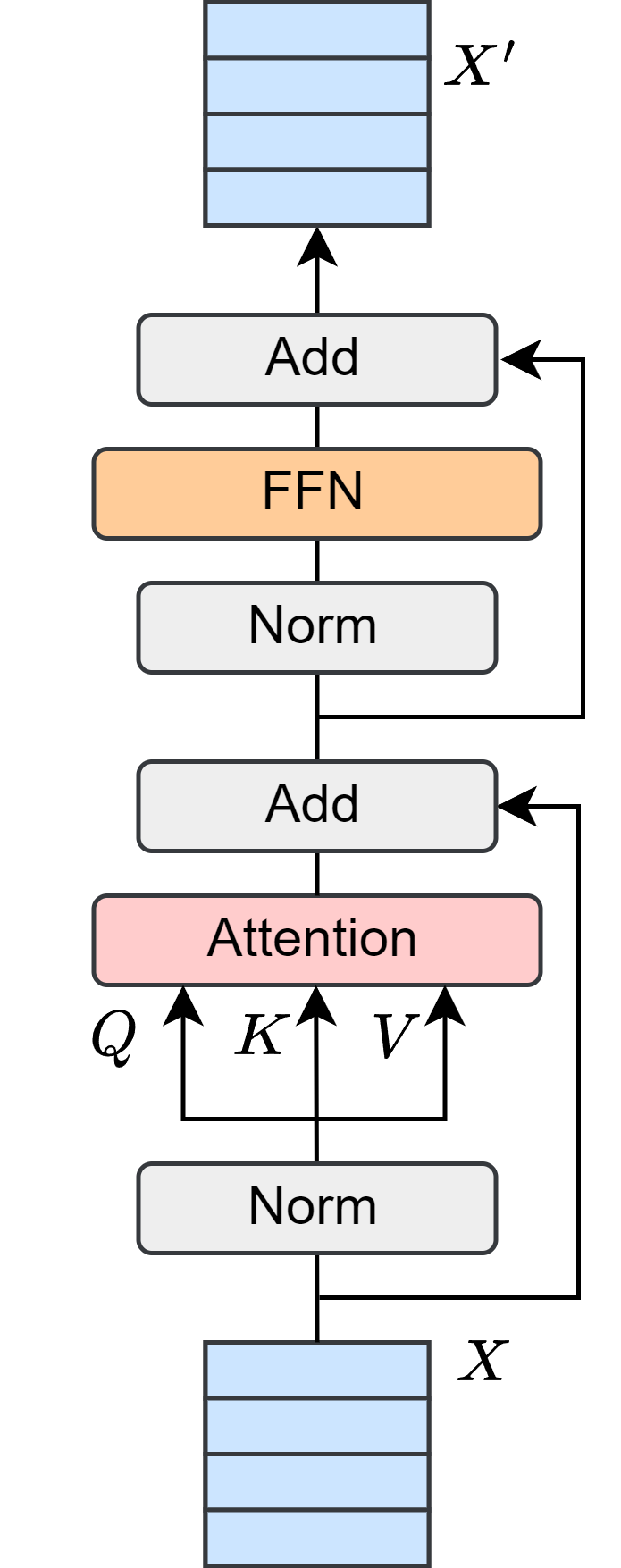}
         \caption{Vanilla Transformer}
        \label{fig:vanilla}
     \end{subfigure}
     \hfill
     \begin{subfigure}[b]{0.32\textwidth}
         \centering
         \includegraphics[width=\textwidth]{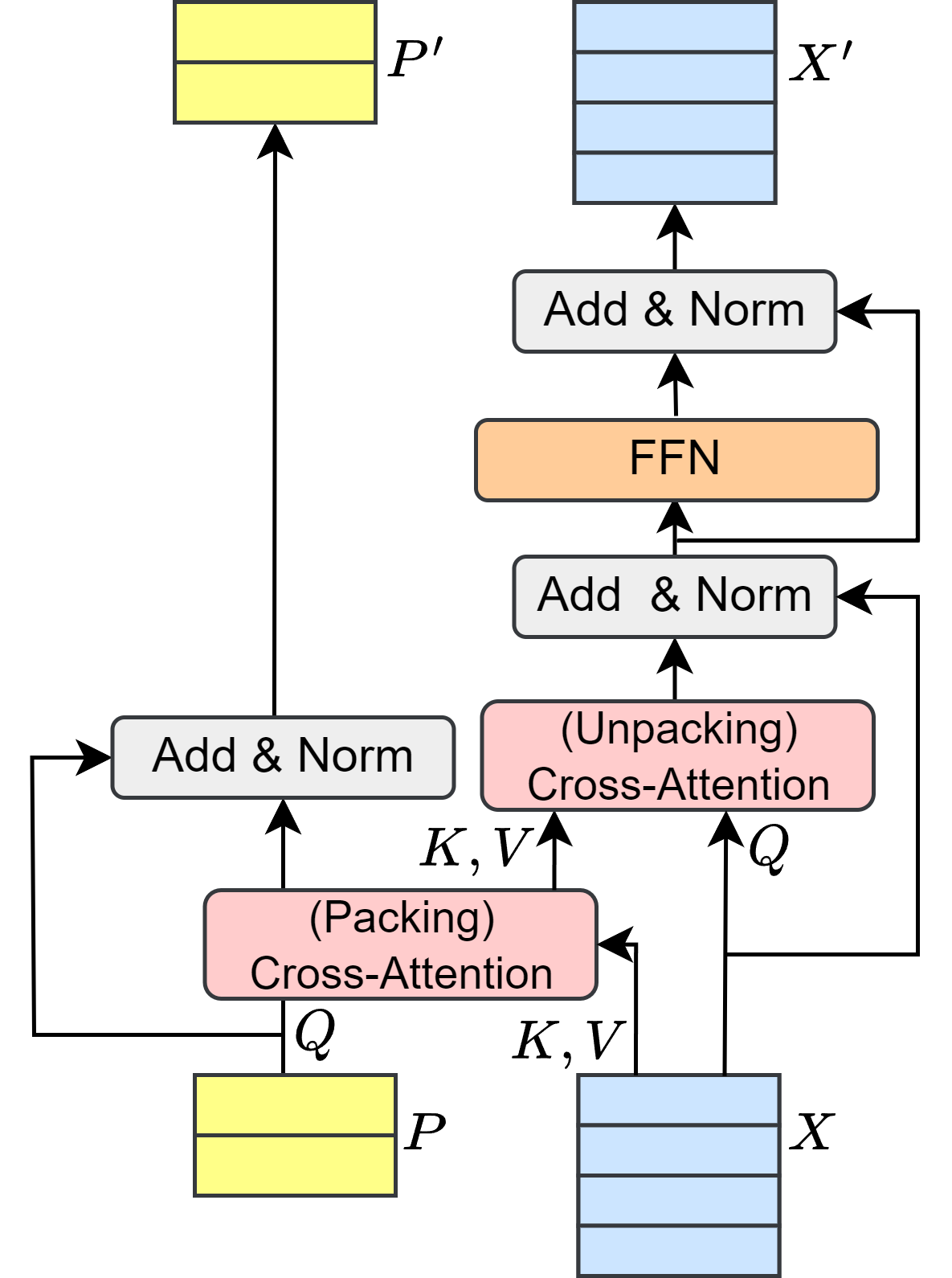}
         \caption{Luna}
         \label{fig:luna}
     \end{subfigure}
     \begin{subfigure}[b]{0.32\textwidth}
         \centering
         \includegraphics[width=0.9\textwidth]{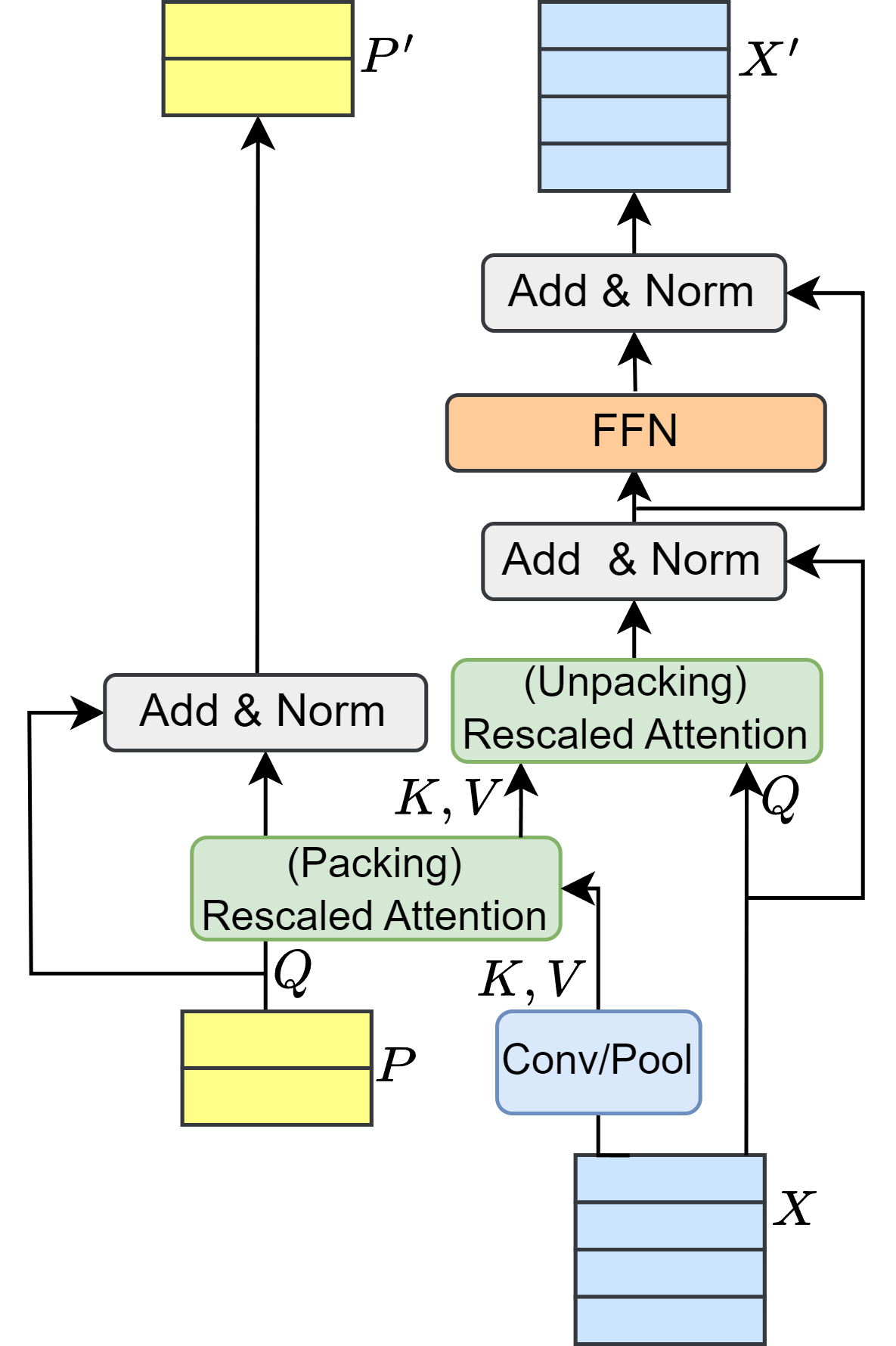}
         \caption{ConvLuna}
         \label{fig:convluna}
     \end{subfigure}
     \caption{Comparison of the ordinary Transformer, Luna and the proposed ConvLuna encoder blocks (we refer to them as modules). ``Packing'' and ``unpacking'' attention layers in Luna are analogous to the attentions in the vanilla Transformer, while the Rescaled Attention additionally multiplies attention logits with a learnable parameter. The ``unpacking" attention of ConvLuna is also denoted as rescalable; however, in our experiments we keep the normalization fixed and identical to the vanilla Transformer.}
     \label{fig:modules}
\end{figure}

The attention \citep{attention-is-all} performs the following calculation over the three matrix inputs $Q, K$ and $V$:

\begin{align*}
\text{MultiHeadAttention}(Q, K, V) &= [\text{head}_1, \text{head}_2, \dots, \text{head}_h]W^O\\
\text{head}_h &= \text{Att}(QW^Q_h, KW^K_h, VW^V_h)
~
\intertext{where}
~
\text{Att}(Q, K, V) &= \text{softmax}\left(\frac{QK^T}{\sqrt{d_h}}\right)V
\end{align*}

and where $W_h$ is the corresponding weight matrix vertical slice for a particular attention head. For simplicity, most of Transformer implementations keep input sizes $Q \in \mathbf{R}^{L \times d}$, $K, V \in \mathbf{R}^{H \times d}$ and weight matrices $W^Q_h, W^K_h, W^V_h, W^O_h \in \mathbf{R}^{d \times d_h}$, $d_h = d/h$. The matrix $\text{softmax}(QK^Td_h^{-1/2}) \in \mathbf{R}^{L \times H}$ is often being referred to as the attention matrix, and may be interpreted as a matrix of relevance scores between $Q$ and $K$ vectors. The Transformer encoder module\footnote{To avoid the overloaded term ``layer", by ``module'' we denote the whole encoder/decoder block, consisting of attentions, FFNs and skip connections, see Figure~\ref{fig:modules}.} equation is thus (see Figure~\ref{fig:vanilla}):
\begin{align*}
    X_{\text{norm}} &= \text{LayerNorm}(X)
    \\
    I &= X + \text{MultiHeadAttention}(X_{\text{norm}}, X_{\text{norm}}, X_{\text{norm}})
    \\
    X' &= I + \text{FFN}(\text{LayerNorm}(I))
\end{align*}

Since the computation of the $L \times H$ attention matrix may be expensive (e.g. quadratic when $Q = K = V$), Luna \citep{luna} factorizes the computation into two attentions (Figure~\ref{fig:luna}): one (``packing'') with $Q$ given as a memory of a fixed length (we denote it as $P$ in the figure) and $K, V$ being module inputs (we denote them as $X$); and the second one (``unpacking'') with the $Q$ given as the module input and $K, V$ as packing attention outputs (Figure~\ref{fig:luna}):
\begin{align*}
    P_{\text{packed}} &= \text{MultiHeadAttention}(P, X, X)
    \\
    X_{\text{unpacked}} &= \text{MultiHeadAttention}(X, P_{\text{packed}}, P_{\text{packed}})
    \\
    I &= \text{LayerNorm}(X + X_{\text{unpacked}})
    \\
    P', X' &= \text{LayerNorm}(P + P_{\text{packed}}), \text{LayerNorm}(\text{FFN}(I) + I)
\end{align*}

The packing attention output serves as the memory input into the next module after applying a skip connection, while the unpacking attention output proceeds to summation with the module input and the residual FFN as in the original Transformer model. Since the $P$ length is constant and typically significantly lower than the length of $X$, the computation of both the packing and unpacking attentions has linear complexity in $H$, which considerably accelerates the computation for longer inputs compared to vanilla Transformer.

%Luna redefines attention processing with its 'Pack and Unpack Attention' strategy \cite{luna}, which splits the conventional attention operation into two nested linear functions that effectively condense and later reconstruct context information using an additional fixed-length input sequence. By outputting this sequence as part of each Luna layer's results and employing shared parameters to manage complexity, the model captures contextual details without a parameter explosion. For scenarios requiring causal attention, Luna incorporates specialized activation functions to prevent future information leakage, ensuring compatibility with tasks like autoregressive decoding. Despite these innovations, the computational demand remains linear, positioning Luna as an efficient alternative to traditional attention in various sequence modeling contexts.

\section{ConvLuna Description}
\label{convluna_desc}

The proposed model differs from the Luna in the packing attention implementation. We change the following:
\begin{itemize}
    \item \textbf{Convolution/MaxPooling layers}, which we apply on keys and values to filter input signal.
    \item \textbf{Learnable softmax temperature}, which we use to rescale attention logits.
\end{itemize}

The main change we introduce into the model is applying either a convolution or pooling on the input in the ``packing" attention. In particular, we apply it on the keys and values (which represent an input text), and not on the queries (which come from the memory); both operations only affect the length dimension, and there is no depthwise interaction in convolutions. The computation flow is similar to the Luna; however, we employ a slightly changed attention module for ``packing" attention:

\begin{align*}
\text{RescaledAttention}(Q, K, V) &= [\text{head}_1, \text{head}_2, \dots, \text{head}_h]W^O\\
\text{head}_h &= \text{RescaledAtt}(QW^Q_h, KW^K_h, VW^V_h)\\
\text{RescaledAtt}(Q, K, V) &= \text{softmax}\left(\frac{Q\tilde{K}^T}{\exp(\tau)}\right)\tilde{V}\\
\tilde{K}, \tilde{V} &= \text{FilterOp}(K), \text{FilterOp}(V)
\end{align*}

where FilterOp is either a convolution or max pooling. The hyperparameters of the FilterOp are kernel size $K$ and stride $S$; given the input sequence of size $L \times d$ (where $L$ is input length and $d$ is the hidden dimension), we apply a kernel of size ($K$, 1) on it with a stride ($S$, 1). We leverage the computer vision interpretation of the operations as filtering the input signal, and leaving only a relevant fraction of it.

Another change we made in the model is the learnable softmax temperature. The original attention formulation (Appendix~\ref{att-luna}) contains the $d^{-\frac{1}{2}}$ multiplier of the query-key product to avoid the softmax saturation; however, during preliminary experiments we have found memory-input product demonstrating less variance than input-input product in the vanilla Transformer. To alleviate this, we have replaced the $d^{-\frac{1}{2}}$ normalizer with $\exp(\tau)$, where $\tau$ is a learnable scalar which we initialize with zero. This does not increase the model flexibility per se, since the term may be absorbed either into $W^Q$ or $W^K$ matrix, but it allows to accelerate the training by increasing the variance of value vectors at the beginning, and imposes only a negligible computational overhead.

\section{Experimental Setup and Hyperparameters}
\label{hyperparams}

We examine our models on the subset of the Long Range Arena (LRA; \citealp{lra}) benchmark, which consists of four classification tasks: %binary sentiment analysis of BPE-encoded texts, BPE-encoded text matching task (predicting the citation link between two documents), ListOps (math operations over nested lists of digits) and CIFAR-10 (an image classification as a sequence of pixels).

\begin{itemize}
    \item \textbf{Byte pair-encoded (BPE) text classification.} This task consists in binary sentiment classification of the IMDB dataset texts encoded as byte pairs. This creates input sequences up to $4k$ tokens long with relatively short subword units.
    \item \textbf{BPE text matching}. The dataset is the ACL Anthology Network \citep{acl}, encoded in a way similar to the previous task with sequences up to $4k$ tokens long. The model needs to process two inputs and to use the concatenated hidden representations as an input to the final layer to classify, whether there is a citation link between the documents.
    \item \textbf{ListOps}. This task consists in processing nested arrays of digits, coupled with aggregation operations such as max, min, median and sum modulo, up to $2k$ tokens long. The model has to predict the correct answer out of ten, and this task tests the ability of the model to process hierarchical inputs. An example input looks like:
    \\
    {\fontfamily{qcr}\selectfont[MED 9 [MAX 4 [MIN 6 3 7 8 9 X 1 2 \dots }
    \item \textbf{CIFAR-10 image classification}. This task is an image classification, but the input is represented as a sequence of pixel values. The data are CIFAR-10 dataset images \citep{cifar}, converted to 8-bit grayscale and flattened inputs 1024 tokens long. 
\end{itemize}

The benchmark suggests training from scratch and limiting the additional parametrization, which allows for a fairer comparison with other models.

Table~\ref{tab:hyper_lra} lists the values of hyperparameters for our experiments. All the ConvLuna models share projection matrices between packing and unpacking attention, and share the $W^K$ and $W^V = \mathbf{I}_d$ weights within a module to increase throughput. We have found the filtering (convolution or pooling) operations hyperparameters relatively shareable across tasks, with only ListOps benefiting from another configuration. We apply both operations length-wise, with the stride being the main hyperparameter to trade the throughput for performance. To better see how memory is being utilized by the model, we use averaged memory cells instead of the [CLS] token as an input embedding before the final layer.

\begin{table}[t]
    \centering
    \resizebox{0.7\columnwidth}{!}{
    \begin{tabular}{l|c c c c }
        \toprule
         & & & \\
         \textbf{Parameter} & \textbf{Classif.} & \textbf{Matching} & \textbf{ListOps} & \textbf{CIFAR} \\
         & & & \\
         \midrule
         Seq. Length & 4000 & 4000 & 2000 & 1024\\
         Batch Size & 32 & 32 & 32 & 64\\
         Training Steps & 25 000 & 30 000 & 50 000 & 30 000\\
         Optimizer & \multicolumn{4}{c}{AdamW ($\beta_1 = 0.9$, $\beta_2 = 0.999$)}\\
         Base LR & 0.005 & 0.015 & 0.005 & 0.01\\
         Weight Decay & 0.01 & 0.04 & 0.01 & 0.01\\
         Warmup Steps & 8000 & 8000 & 1000 & 700\\
         Schedule & \multicolumn{4}{c}{Base LR * Warmup * Sqrt Decay}\\
         Warmup Mul. & \multicolumn{4}{c}{$min(1, \text{Current Step} / \text{Warmup Steps})$}\\
         Sqrt Decay Mul. & \multicolumn{4}{c}{$1 / \sqrt{max(\text{CurrentStep}, \text{WarmupSteps})}$}\\
         Loss & \multicolumn{4}{c}{CCE}\\
         Blocks & 4 & 4 & 6 & 8\\
         Heads & \multicolumn{4}{c}{4}\\
         Hidden dim. & \multicolumn{4}{c}{128}\\
         QKV dim. & \multicolumn{4}{c}{128}\\
         MLP dim. & \multicolumn{4}{c}{512}\\
         Filter Op. & MaxPool & MaxPool & Conv & MaxPool\\
         Kernel Size & 4 & 4 & 32 & 4\\
         Stride & \multicolumn{4}{c}{1}\\
         Dropout & 0.1 & 0.1 & 0.1 & 0.0\\
         Activation & \multicolumn{4}{c}{GELU}\\
         Pooling & \multicolumn{4}{c}{Memory Cells Average}\\
         Pos. encoding & \multicolumn{4}{c}{Learnable Absolute}\\
         \bottomrule
    \end{tabular}
    }
    \caption{Hyperparameters used for the LRA experiments.}
    \label{tab:hyper_lra}
\end{table}

\section{Ablation Study}
\label{ablation}

\begin{table}[bt]
    \centering
    \begin{center}
    \resizebox{0.9\columnwidth}{!}{
    \begin{tabular}{llllll}
    \toprule
   Model &  Classification $\uparrow$ & Matching $\uparrow$ & ListOps $\uparrow$ & CIFAR-10 $\uparrow$ & Average $\uparrow$ \\
   % \\
    \midrule
    OnlyScaling-1 & 65.45 $\pm$ 0.30 & 72.46 $\pm$ 0.68 & 37.12 $\pm$ 0.47 & 49.23 $\pm$ 1.35 & 56.07 \\
    OnlyScaling-16 & 66.81 $\pm$ 0.21 & 74.99 $\pm$ 1.67 & 37.08 $\pm$ 0.18 & 48.67 $\pm$ 0.65 & 56.89 \\
    OnlyScaling-256 & 66.95 $\pm$ 0.20 & 75.55 $\pm$ 0.61 & 37.52 $\pm$ 0.33 & 48.90 $\pm$ 1.01 & 57.23 \\
    \midrule
    OnlyFiltering-1 & \textbf{82.39} $\pm$ 0.62 & 82.38 $\pm$ 0.73 & 40.44 $\pm$ 2.31 & \textbf{58.69} $\pm$ 1.13 & \textbf{65.98} \\
    OnlyFiltering-16 & 80.89 $\pm$ 0.44 & \textbf{82.66} $\pm$ 1.46 & \textbf{41.48} $\pm$ 1.74 & 58.06 $\pm$ 0.59 & 65.77 \\
    OnlyFiltering-256 & 80.27 $\pm$ 0.97 & 82.10 $\pm$ 1.68 & 41.24 $\pm$ 1.42 & 58.12 $\pm$ 0.66 & 65.43 \\
    \bottomrule
    \end{tabular}
    }
    \end{center}
    \caption{Results of the ablation experiments. OnlyScaling denotes the setup where we apply no filtering operations in packing attention, while OnlyFiltering means that we do not use the learnable softmax temperature there. Following the Table~\ref{tab:lra-results}, we report accuracy score mean and standard deviation across five runs for each setup, and highlight the best result for each task with bold.}
    \label{tab:ablation}
\end{table}

During the ablation experiments, we run the training of our models again, but now applying only one change at a time. The Table~\ref{tab:ablation} reveals that the applying convolution or pooling indeed leads to significant performance improvements on the considered tasks. However, a straighforward improvement is not the case when we apply the learnable logit scaling; in some setups it can improve the accuracy on its own or in combination with the convolution/pooling, but in some setups (CIFAR-10) it degrades the performance.

Although the effect of the learnable logit scaling is not the same for all the considered tasks, the combination of both methods yields the best average accuracy across the all setups (Table~\ref{tab:lra-results}). We also refer to the fact that the initial value of the scalar is lower than the fixed $d^{-\frac{1}{2}}$ of the vanilla attention, which in some cases may be the more important factor than the trainable parameter itself. Dividing the attention logits with a lower value leads to more saturated and varied attention scores at the training start compared to the nearly uniform distribution of the vanilla attention, which results in a better result and a faster convergence in some tasks.

We provide the following interpretation: for the setups such as ListOps, where the models demonstrate very slow convergence, or Classification, where we hypothesize that the model operates similarly to the Bag-of-Words and thus needs to focus more on particular tokens, such behavior that forces more attention to a smaller subset of tokens may be beneficial. At the same time, within the CIFAR-10 and Matching tasks the answer is unlikely to be dependent only on a small subset of input tokens, so a more focused attention at the beginning stops being a benefit and the additional parameter may contribute to the overfitting.

\end{document}